\documentclass{article}

\usepackage{arxiv}

\usepackage{amsfonts}
\usepackage{amstext}
\usepackage{amsmath}
\usepackage{mathtools}

\usepackage{graphicx}

\DeclareMathOperator*{\argmax}{arg\,max}
\DeclareMathOperator*{\argmin}{arg\,min}

\begin{document}

\title{Security Matters: A Survey on Adversarial Machine Learning}

\author{
 Guofu Li \\
  College of Communication and Art Design \\
  University of Shanghai for Science and Technology \\
  Shanghai, China \\
  \texttt{li.guofu.l@gmail.com} \\
 \And
 Pengjia Zhu \\
  State Street Corporation \\ 
  Hangzhou, China \\
  \texttt{zhupengjia@gmail.com}
 \And
 Jin Li \\
  School of Computer Science and Educational Software \\
  Guangzhou University \\
  Guangzhou, China \\
  \texttt{jinli71@gmail.com}
 \And
 Zhemin Yang \\
  School of Computer Science \\
  Fudan University \\
  Shanghai, China \\
  \texttt{yangzhemin@fudan.edu.cn}
 \And
 Ning Cao \\
  College of Information Engineering \\
  Qingdao Binhai University \\ 
  Qingdao, China \\
  \texttt{ning.cao2008@hotmail.com}
 \And
 Zhiyi Chen \\
  College of Communication and Art Design \\
  University of Shanghai for Science and Technology \\
  Shanghai, China \\
  \texttt{iamedithchen@gmail.com}
}

\date{}
\maketitle

\begin{abstract}
	Adversarial machine learning is a fast growing research area, which considers the scenarios when machine learning systems may face potential adversarial attackers, who intentionally synthesize input data to make a well-trained model to make mistake. 
	It always involves a defending side, usually a classifier, and an attacking side that aims to cause incorrect output. 
	The earliest studies on the adversarial examples for machine learning algorithms start from the information security area, which considers a much wider varieties of attacking methods. 
	But recent research focus that popularized by the deep learning community places strong emphasis on how the ``imperceivable'' perturbations on the normal inputs may cause dramatic mistakes by the deep learning with supposed super-human accuracy. 
	This paper serves to give a comprehensive introduction to a range of aspects of the adversarial deep learning topic, including its foundations, typical attacking and defending strategies, and some extended studies. 
%	We also share our points of view on the root cause of its existence and possible future directions of this research field. 

\keywords{
Adversarial \and Security \and Deep Learning.
}

\end{abstract}

%\tableofcontents

\section{Introduction}
\label{sec:intro}

	The key difference between machine learning and pure optimization is that, machine learning algorithm is expected to have the ability to generalize from the given (training) dataset to the unseen (test) dataset. Though there can be a variety of the ways how the data are collected and/or annotated, one underly assumption is shared by most of the scenarios, that there is always a stationary stochastic process that generates each and every sample point in both of the dataset. In other word, both samples in either datasets follow the independent and identical distribution ({\em i.i.d}) promise. But in many real-world settings, this assumption maybe over-simplified. 

	Early researches in the machine learning community realize this issue when they try to measure and improve the {\em robustness} of the classifiers, in the sense that it should ties it best to be immune to the anomaly points in the training data or the test data. 
	Studies on this topic can trace back to Wald's robustness of statistical decision theory \cite{wald1945statistical}. Researches fall in this category commonly use a term ``mini-max'' robustness with somehow similar intuition to the modern ``adversarial'' description of the task. 
	For instance, a research conducted by Lanckriet et al. \cite{lanckriet2003a} defines a task of finding the optimal worst-case (thus mini-max) misclassification probability on a binary classification problem on all possible choices of class-conditional densities, but with a fixed known the mean and covariance matrix, and is then reduced to a convex second order cone optimization problem. 
	% After translating the problem into a convex second order cone optimization problem, they prove that the optimal decision hyperplane can be derived by minimizes the maximum of the {\em Mahalanobis} distances to the two classes, and on the common tangent of both of the ellipsoids. They also exploit the Mercer kernel to generalize the algorithm to nonlinear classifiers. 
	% may cause a significant degeneration of classifier's performance, if these points lies in the critical boundaries of the decision \cite{}. 

	% There is one important source of this ``anomaly'' that are different in nature to what the previous researches focus on, which may result to unexpected damages to an online working system. 
	Another story-line starts from the information security area, which the mini-max games are ubiquitous.
	For instance, when a security controller designed to prevent the online service from malicious requests is ensured by a binary classifier, the requests generated by the malicious users will mostly likely not follow the ``stationary'' assumption. Anomalies of this type may occur only in the test dataset, commonly not due to the ``rare events'', but attribute to the intentional attacks on the system launched by an ``adversarial'' party. 
	Besides the adversarial during test time, research by Biggio et al. also show that a few well-selected training instances may cause severe poisoning attacks against Support Vector Machines (SVM), together with a new principled defense method highly resilient against all poisoning attacks \cite{biggio2012poisoning}. 
	% extra reference work: Manipulating Machine Learning: Poisoning Attacks and Countermeasures for Regression Learning
	More recently, Jagielski et al.	studies exclusively on the poisoning attack on linear regression, and propose a theoretically-grounded optimization framework \cite{jagielski2018manipulating}. 
	In sensitive scenarios such as malware classification \cite{grosse2016adversarial}, 
	%!! should rephrase this sentence ...
	the existence of adversarial examples has put forward great challenges on the applicability of machine learning based approaches \cite{dahl2013large, grosse2016adversarial}: 
	The system security can be severely compromised, if an adversary is able to make slight modifications on a malware file to make it remain a malware, but is thus misclassified as benign by the classifier. 
	These potential security breaches may become even more sensitive to {\em safety-critical} tasks such as an airborne collision avoidance system for unmanned aircraft \cite{katz2017reluplex} or self-driving car \cite{DBLP:conf/sosp/PeiCYJ17}. 

	The specialness within the term ``adversarial machine learning'' is the involvement of two ``adversarial'' parties within an mini-max game, in which the attacker side learns to produce the adversarial examples to trick the defender (i.e., the classifier induced by a machine learning algorithm), while the classifier tries to defend attacks of this kind. 
	Lowd and Meek \cite{lowd2005adversarial} were among the first who employ the term ``adversarial'', and introduce the concept of ``adversarial learning'' to refer to a learning problem of Adversarial Classifier Reverse Engineering (ACRE), in which the algorithms aims to collect sufficient information about a classifier to construct adversarial attacks. 
	
	In contrary to Lowd and Meek's task definition, Huang et al., \cite{huang2011adversarial} use the term ``adversarial machine learning'', which refers to the study of effective machine learning techniques against an adversarial opponent. Although the task definition of the two seem to be rather opposite, their ultimate goals are however coincident. From the point view of the attack side, {\em adversarial} means one needs to base the attack activity on the knowledge or heuristics of the defense (classifier) side instead of random moves, while the defense side also need to improve the robustness on the ``worst cases'' instead of i.i.d inputs. 	
   Hereby, this paper serves to present a comprehensive review on the literatures over the adversarial problem in machine learning, mainly focuses on the attacking and defending strategies in deep neural networks.

\section{Backgrounds}

	\subsection{The ``Imperceivable'' Perturbation}
	\label{sec:imp-pert}

	In Section \ref{sec:intro}, we introduced the concept of ``adversarial learning'' as a mini-max game\cite{bruckner2009nash}, in which the attacker aims to maximize its likelihood to trick the classifier, while the classifier aims to minimize its risk in such worst case. 
	There are many ways to launch attacks on a machine learning based system. One could inject malicious data points into the training data, to manipulate the training procedure, or one can use the model information (e.g., the gradient w.r.t the input) to construct the adversarial. Attackers of any types above, together with its counter party naturally involves a pair of ``adversarial'' parties, and thus can be categorized into the ``adversarial machine learning''. 
	
	However, the recent researches in the deep learning community impose more restriction on the definition of ``adversarial''. It commonly refers to a process of producing ``small and imperceptible perturbations'' (or distortions in some literature) to human to the normal example, which however, can fool the machine learning algorithms. 
	Formally speaking, a classification function $F : \mathbb{R}^m  \to {1 \dots k}$ maps the input real-valued vectors to a discrete label set. For deep neural networks, the classifier's output is commonly based on a probability distribution produced by $f$ that gives a probability distribution on the class labels by a final softmax layer. 
	So, for a given input instance $\boldsymbol{x}$, the output class of a neural network can be written as $F(\boldsymbol{x}) = \argmax_j f_j(\boldsymbol{x})$. 
	Thus, the goal of the attacker is to find the adversarial input $\boldsymbol{x}'$ that is close to $\boldsymbol{x}$ such that $F(\boldsymbol{x}') \neq F(\boldsymbol{x})$. 

	The parameters of the network $\boldsymbol{\theta}$ is then optimized by minimizing the expectation of a loss function $\mathbb{E}_{\boldsymbol{x} \sim \text{data}} \left[ \mathrm{loss}_{f, l} (\boldsymbol{x}) \right] $, where $f$ outputs the vector form of a categorical distribution $q$, such that $q_i = \mathrm{softmax}(i)$, 
	and $l$ is the expected output class label for the input $\boldsymbol{x}$, commonly represented as a vector form of the discrete Dirac distribution $p$, such that
	\begin{equation}
		p_i = 
		\begin{cases}
			1 	& \quad \text{if } i = l \\
			0	& \quad \text{otherwise .}
		\end{cases}
	\end{equation}
	\label{eqn:dirac}
	For classification tasks, the loss function usually appears in the form of cross entropy $H(p, q) = \mathbb{E}_{p} [ - \log q ]$, which is equivalent to the Kullback-Leibler divergence of $q$ from $p$ for a fixed $p$.  
	
	The origin of the study on adversarial in deep learning may trace back to the profound discovery made by Szegedy et al. \cite{szegedy2013intriguing}, pointing out that in many of the deep learning based computer vision systems, a imperceptibly small perturbation on the input image may cause a total and seemingly uninterpretable failure in the classification result. This discovery enforces the researchers to rethink the question of ``What went wrong?'', despite the great success of the ``super-human'' performance in many of the computer vision tasks using deep Convolutional Neural Networks (CNNs). As such, there are several follow-up questions:
	\begin{enumerate}
		\item Why these perturbations on the input are imperceptible to human but have large impact on systems with ``super-human'' performance?
		\item How can adversarial examples of this type can be constructed? 
		\item How do we prevent from the classifiers to make mistakes on these adversarial examples? 
	\end{enumerate}

	\subsection{Measuring the Perturbation}
	
	\subsubsection{The $\ell^{p}$ Metric}
	
	To have a more solid foundation to study this problem, we need a more precise way to measure the perturbation. Although the ways to define ``imperceptible'' are task dependent, most of which are based on the $\ell^p$ norm distance. The $\ell^p$ norm distance between two points $\boldsymbol{x}$ and $\boldsymbol{x'}$ in a high-dimension space can be written as $\|\boldsymbol{x} - \boldsymbol{x}'\|_p$ , where the operator p-norm $\| \cdot \|_p$ on an $n$-dimensional vector is defined as
	\begin{equation}
		\| \boldsymbol{v} \| = \left( \sum_{i=1}^{n} \| v_i \|^{p} \right) ^\frac{1}{p} \text{ .}
	\end{equation}
	The value of $p$ is parameter that may vary due to the need of a specific task. When $p = 2$, the $\ell^2$ distance between two points $\boldsymbol{x}$ and $\boldsymbol{x}'$ (written as $\|\boldsymbol{x} - \boldsymbol{x}'\|_2$) is simply measures the Euclidean distance between $x$ and $x'$, which turns out to be a default choice in many early researches. When $p=1$, the $\|\boldsymbol{x} - \boldsymbol{x}' \|_1$ distance is equivalent to the sum of the absolute value of each dimension, which is also known as the {\em Manhattan distance}. Two special value of $p$ is $0$ and $\infty$: The $\ell^0$ distance between two points measures the number of dimensions that have different values $\sum_{i} \mathbf{1}[ x_i \neq x'_i ]$, 
	and the $\ell^{\infty}$ distance is a conventional way to denote the maximum absolute difference among all dimensions $\max_i \| {x}_i - {x}'_i \|$, where $\mathbf{1}[.]$ is the indicator function, and $x_i$ denotes the value of the i-th dimension of $\boldsymbol{x}$. 

	When the $\ell^2$ norm is used, the distance between two points will accumulate along the each of the dimensions, making the number of dimensions of the input data can lead to significant difference concerning the ``imperceptible''. A simple way to remove the bias due to the dimension number of the input (e.g., the number of pixels of the input image) is by considering the average distortion on a per-dimension basis by 
	$\frac{1}{n} \|\boldsymbol{x} - \boldsymbol{x}'\|_2$, where $n$ is the number of input dimensions. 
	However, this measurement may under-estimate the distortion if $n$ is large. Therefore, in the computer vision scenarios, experiment may often conducted on images of different sizes or resolutions, therefore, $\ell^{\infty}$ is often a preferred choice, which simply measure the maximum change of any of the pixel w.r.t the original input, thus invariable to the image resolution. That is, the distortion for any of the pixel has to be under a fixed bound. 
	
	The actual choice of the distortion measurement can be task-dependent. For example, naive Bayesian classifier is a common choice by many of the natural language processing tasks, which uses boolean features (i.e., if a certain word occurs or not, e.g., as appeared in Lowd and Meek \cite{lowd2005good}), making other measurements may seem more plausible. An alternative choice that show interesting result is the ``one pixel attack'' proposed by Su et al. \cite{su2017one}, or a few pixels proposed by Narodytska and Kasiviswanathan \cite{narodytska2017simple}
	, in which the constrains is imposed on the number of pixels that is modified and is equivalent to the $\ell^0$ norm. 
	One major reason for choosing the $\ell^{\infty}$ measurement by Goodfellow et al. \cite{goodfellow2014explaining} is because they consider the success of the adversarial attacks is due to the accumulated perturbation in a large number of dimensions. But since the $\ell^0$ metric has constraint the number of dimensions to be modified, their proposed explanation is then proved to be unnecessary. 
	
	\subsubsection{Other Metrics}

	Ideally, the reported amount of perturbation due to some measurement should reflect the perceptual impact it has on human's cognition. 
	So in Lowd and Meek's framework \cite{lowd2005adversarial}, they introduce {\em cost function} as a more general form to measure the perturbation, by considering its negative impact on its attacking goal. Opposite to today's common setting, the construction of the adversarial input in their work starts from the malicious sample, which is then modified to evade being detected by the classifier. 
	Thus, two spam emails may both trick a classifier to be a normal mail though, one of them may be effective for its goal, while the other may lose too much of its original. 
	A more recent version is given by Madry et al.\cite{madry2017towards}, which formalize it as a mini-max game, in which an unified cost is defined, and the attacker's goal is to maximize the cost, while the classifier would need to minimize the cost. This point of view can be traced back to the early theoretical work on statistical optimization problem by Wald \cite{wald1945statistical}. 
	
	Although these more generic definitions on the perturbation metrics, the $\ell^p$ norm distance provides an simple and elegant way to calculate this quantity, making it more practical to be embedded inside an optimization problem,. 
	However, one potential issue with the $\ell^p$ norm based perturbation measurement is that it is commonly applied directly on the input space $\mathcal{X}$, the distance in which may not reflect the affect at the perceptual level. This lead to the idea of measuring the distance between the two points after mapped into the latent space $\boldsymbol{z}$ from both the attacking side and the defending side. We will elaborate strategies fall inside this type in Section \ref{sec:gan-attack} and \ref{sec:gan-defend}.
	
	\subsection{Non-deep-learning Adversarial} 
	
	%The robustness of statistic decision functions has been a research area by itself even before the birth of the modern computer \cite{wald1945statistical}, which has a strong connection with the concept of ``adversarial'' in the context of machine learning. 
	%In fact, the research on the adversarial attack and defense of learning is largely depending on the advancement of machine learning algorithm itself. 
	Early research that is similar to the modern concept of ``adversarial'' can be traced back to the work by Dalvi et al. \cite{dalvi2004adversarial}. 
	There have been attempts to build attacks on security systems that built on machine learning algorithms though, they notice that previously approaches to address this problem is by repeated, manual, ad hoc reconstruction of the classifier, which motivates them they develop a formal framework and algorithms for this problem. 
	Similar to the study on the robustness of statistic, they model the classification scene in a security system as a ``mini-max game'' between the classifier and the adversary, in which the classifier should be optimized in a way that has the best performance given the adversary's optimal strategy. 
	Their experiment is exclusively focused on building adversary-aware Naive Bayes classifier \cite{domingos1997optimality} applied on to be made spam detection system. 
	Soon after, a similar idea is presented by Lowd and Meek, who propose the {\em good word attack} \cite{lowd2005good} that  a spammer can transform spam message into legitimate email by simply inserting or appending ``good words''. 
	Wang et al. \cite{wang2018stealing} recently propose a new type of attacking task, namely the {\em hyperparameter stealing attack} task, which aims to identify the hyper-parameter $\lambda$, which balances the training data optimization and the overfitting penalty. 

	One important class of approaches to solving this problem is suggested by Globerson et al. \cite{globerson2006nightmare}, who argues that the key factor to prevent classifier from such plight is to avoid assigning too much weight to any single input feature. The same idea is shared by many of the follow-up researches. For instance, Zhang et al. \cite{zhang2016adversarial} argue that the common practice of feature-reduction process adopted by most machine learning systems turns out to be hazardous for classifier's robustness, and proposed an adversary-aware feature selection method to improve classifier security against evasion attacks. Following the same intuition, Ning et al. \cite{cao2018handling} suggest that using balanced decision tree to build random forests can significantly improve the robustness of the ensemble classifier. 
	
	A recent study by Papernot et al. show that the transferability of adversarial example may even exist among different learning algorithms \cite{papernot2016transferability}, despite the fact that they have different assumptions and intuitions in nature (e.g., decision tree, support vector machine, or deep neural networks, etc.), which gives rise to one type of black-box attack, and will be discussed in Section \ref{sec:black-box}. 
	
	\subsection{Derived Topics}

	\subsubsection{Generative adversarial networks}
	\label{sec:gan}
	
	So far, our discussion on the adversarial learning focuses on a class of machine learning algorithms called {\em discriminative model}, whereas the other side of it is what known as the {\em generative model}. A discriminative model aims to predict the label $y$ of some latent variable based on the observed variable $\boldsymbol{x}$, also known as the {\em classification} task, commonly approached by modeling the conditional probability of $p(y | \boldsymbol{x})$. Comparatively, a generative model is a model that can generate new sample points from the distribution $p(\boldsymbol{x})$. Though originated from the discriminative model, the research on the mini-max game between the two adversarial parties lead to one of the most important advancement in learning generative models, namely the Generative Adversarial Networks (GANs) \cite{goodfellow2014generative}. 
	
	In the setting of adversarial machine learning, the two adversarial parties happen to fit the two classes of the machine learning algorithms: the defending side aims to correctly classify all input examples, regardless of whether it is a naturally occurred or adversarially manipulated; on the other hand, the attacking side aims to {\em generate} new examples that can lead to the mistakes made by the classifier. In adversarial learning, the ultimate goal is to improve the robustness of the classifier by putting it under the strongest adversarial generator. Contrarily, the ultimate goal of the GAN is to find the optimal generator, whose output can not be distinguished from the normal data even by the strongest discriminator. 
	
	In GAN, a generator $\mathcal{G}$ parameterized by $\boldsymbol{\theta}^{(\mathcal{G})}$, aiming to fake realistic samples that cannot be distinguished from real ones, and a discriminator $\mathcal{D}$ parameterized by $\boldsymbol{\theta}^{(\mathcal{D})}$, aiming to detect if an input sample is a genuine one or faked by the generator $\mathcal{G}$, form a pair of adversarial parties.
	The basic version of this mini-max game is a zero-sum game, in which the objective function of the generator $J^{(\mathcal{G})}$ and the objective function of the discriminator $J^{(\mathcal{D})}$ add up to $0$. Under this setting, the game can be expressed in a single value function for the discriminator:
	\begin{equation}
		V\left( \boldsymbol{\theta}^{(\mathcal{D})}, \boldsymbol{\theta}^{(\mathcal{G})} \right) = - J ^{(\mathcal{D})} \left( \boldsymbol{\theta}^{(\mathcal{D})}, \boldsymbol{\theta}^{(\mathcal{G})} \right) \text{ ,}
	\end{equation}
	in which the generator $\mathcal{G}$ produces faked samples from the latent codes $\boldsymbol{z}$, and the discriminator $\mathcal{D}$ gives the probability of an observed sample being genuine. Since the goal of the mini-max game is to find the optimal generator $\mathcal{G}$ which can draw new example from the distribution of $p_{\text{data}}$, the optimization objective can be written as
	\begin{equation}
		\boldsymbol{\theta}^{(\mathcal{G})*} = \argmin_{\boldsymbol{\theta}^{(\mathcal{G})}} \max_{\boldsymbol{\theta}^{(\mathcal{D})}} V\left( \boldsymbol{\theta}^{(\mathcal{D})}, \boldsymbol{\theta}^{(\mathcal{G})} \right) \text{.}
		\label{eq:gan-value}
	\end{equation}
	The original proposed objective function for the discriminator is as follows:
	\begin{equation}
	\begin{split}
		J^{(\mathcal{D})}\left( \boldsymbol{\theta}^{(\mathcal{D})}, \boldsymbol{\theta}^{(\mathcal{G})} \right) = & -\frac{1}{2} \mathbb{E}_{\boldsymbol{x} \sim p_{\mathrm{data}}} \log \mathcal{D}(\boldsymbol{x}) \\
		& - \frac{1}{2} \mathbb{E}_{\boldsymbol{z}} \log (1 - \mathcal{D} (\mathcal{G}(\boldsymbol{z}))) \text{ .}
	\end{split}
	\end{equation}
	Following Equation \ref{eq:gan-value}, we can derive the objective function for the generator $\mathcal{G}$ as
	\begin{equation}
		J^{(\mathcal{G})} = - \frac{1}{2} \mathbb{E}_{\boldsymbol{z}} \log \mathcal{D}(\mathcal{G}(\boldsymbol{z})) \text{ ,}
	\end{equation}
	by dropping the constant term in the $J^{(\mathcal{D})}$ with respective to $\mathcal{G}$. 
	
	The huge success of GAN in synthesizing realistic images from a low dimension input $\boldsymbol{z}$ draws exponential growth of research focus. Soon, the original form the GAN is found to suffer several problems, most importantly:
	\begin{enumerate}
		\item The training process can be extremely unstable, and often fail to converge at all;
		\item The quality of the update gradient received by the generator drops as the discriminator improves. 
	\end{enumerate}
	Many of the extended researches are devoted to how to stabilize the training process by studying different possibilities of the objective functions. One of the most important improvements over the GAN since its invention is the Wasserstein GAN \cite{arjovsky2017wasserstein, NIPS2017_7159}, which uses the Wasserstein distance (a.k.a., the earth moving distance) to replace the original Kullback-Leibler divergence based distance. Currently, a large portion of the state-of-art research on deep learning is moving towards the GAN based approaches \cite{Creswell2017Generative}. 
	
	Although {\em discriminative model} and {\em generative model} are the two faces of machine learning, they are not parallel to each other. For instance in GAN, a discriminative model is introduced to find the high-quality update gradient for optimizing the generative model. It is also true for adversarial machine learning, in which a generator that captures the data distribution well can be useful to improve the robustness of the classifier, as we will discuss in Section \ref{sec:pert-recovery}. Moreover, the GAN model can be directly involved in both the attacking side and the defending side. 
	
	\subsubsection{Adversarial Reprogramming}
	
	Deep learning is a type of representation learning, whose goal is to find the best representation transformations via a hierarchy of levels, such that the data in a certain domain that best fit the task goal. Previous researches show that many of the low-level features (e.g., edge detection) can be shared among a wide variety of tasks, and can considerably leverage the data from a task with rich data supply to a task with few, leading to the a special type of {\em transfer learning}. 
	When employing the transfer learning method, one would reuse the representation transformation learnt from another task, and then {\em re-purposing} the model by re-training the parameters (possibly with extra layers).

	A recent research work by F. Elsayed et al. \cite{elsayed2018adversarial} combine the idea of transfer learning with the adversarial behavior of the deep neural networks, 
	seeking to achieve the ``re-purposing'' not by re-training the model, but by add specially designed perturbations to the inputs. 
	They propose a new {\em reprogramming task}, in which the target model is reprogrammed to perform a new task chosen by the attacker,
	by using a single unrestricted magnitude adversarial perturbation across all natural inputs to the machine learning model,
	even if the model was not originally trained for this task. 
	These perturbations can be considered as a special type of program, and feeding inputs with these them are identical to {\em reprogramming} the neural network. 
	
	In adversarial reprogramming, it is important to distinguish between the two tasks:
	\begin{enumerate}
		\item the original task $T$, which consists of the input $\boldsymbol{x} \in \mathbb{R}^{n \times n \times 3}$ (suppose the input is an image of size $n \times n$ with 3 channels) and the output label $y$, 
		\item and the adversarial (or more precisely, the reprogramming) task $T'$, which consists of the input $\boldsymbol{\tilde{x}} \in \mathbb{R}^{k \times k \times 3}$ with $k < n$, such that the input for $T'$ to be smaller in size than that for $T$, and the desired output $y_\text{adv}$. 
	\end{enumerate}

	As such, the perturbation in this scenario $\boldsymbol{W} \in \mathbb{R}^{n \times n \times 3}$ is designed for the input of the adversarial task $\tilde{\boldsymbol{x}}$, which is firstly padded into $\tilde{\boldsymbol{X}} \in \mathbb{R}^{n \times n \times 3}$ to fit the input shape of the original network. The extra space of the input allows us to create an optional fixed mask $\boldsymbol{M} \in \{0, 1\}^{n \times n \times 3}$ to make the perturbed input 
	\begin{equation}
	\boldsymbol{X}_\text{adv} = \boldsymbol{X} + \tanh (\boldsymbol{W} \odot \boldsymbol{M}) \text{ ,}
	\end{equation}
	such that the original input $\tilde{\boldsymbol{x}}$ may stay visually intact for easier inspection. Here, $\tanh(\cdot)$ ensures the perturbation is valid in range, and $\odot$ denotes the element-wise product. 
	In addition, the number of possible output labels $y_\text{adv}$ for the adversarial task may be fewer than that for the original task, which requires us to create a mapping function $h_g$, possibly by hard-coding manually. For instance, if we reprogram the network originally trained for ImageNet task, which has 1,000 output labels, for a MNIST task, which only has 10 output labels, then we need to map averagely 100 values of $y$ into one value of $y_\text{adv}$. 
	Hence, the adversarial reprogramming task can be formulated as follows:
	\begin{equation}
		\hat{\boldsymbol{W}} = \argmin_{\boldsymbol{W}} {(-\log P(h_g(y_\text{adv}) | \boldsymbol{X}_\text{adv}) + \lambda \| \boldsymbol{W} \|_2^2)} \text{, }
	\end{equation}
	where $\boldsymbol{W}$ is the adversarial program parameters, and $\lambda$ is a hyper-parameter to scale the weight penalty to reduce overfitting. 
	
	Their empirical study show that trained neural networks are more susceptible to adversarial reprogramming than random networks, which is a strong evidence to support the hypothesis that adversarial program serves as a special type of transfer learning to repurpose a well-trained network. 
	Their findings also lead to a wide range of potential applications, and raise even more questions about the network security to be explored. For instance, it is possible for a malicious user to abuse the image classifier hosted on cloud to solve image captchas. 
	
\section{Attacking Strategies}
	\label{sec:attack}

	\subsection{A Taxonomy of Attacks}
	
	There are more than one way to attack a system. Defining the {\em threat model}, a concept brought from the information security community, which specifies what the attacker wants to achieve, what kind of knowledge she possesses, and what steps she would follow, allows us to analyze the possible attacks systematically and theoretically. 
	Barreno et al. \cite{barreno2006can} propose a classification scheme of the strategies, which consists of three dimensions. 

	\begin{itemize}
		\item {\bf Influence}
		In this dimension, one would need to consider how much control the attack has on the training process of the machine learning algorithm. In some occasions, the attacker has the ability to influence the training data during the training process, which is called {\em causative} attack. One typical class of of the causative attack is known as the poisoning attack, in which poisonous inputs are injected into the training data to alter the fitted decision boundary \cite{biggio2012poisoning, jagielski2018manipulating}. 
		% extra reference: Trojaning Attack on Neural Networks
		The trojan attack proposed by Liu et al. \cite{liu2018trojaning} is another example of causative attack, which instead of tampering with the original training process, produces a new model and an {\em attack trigger} or {\em trojan trigger} by adding extra training data to retrain the original model. 
		In most real-life scenarios, the attacker may only be able to {\em probe} and {\em explore} the behavior of the classifier, and infer the underly structure of it, but have no control over the training process. Attacks of this kind are called {\em exploratory} attack. 

	    \item {\bf Specificity}
	    This dimension determines how much {\em specificity} the attacker's goal is set. At one end of the dimension is the {\em targeted} attack, in which the one specific input point or a small set of input points is considered valuable to the attacker. At the other end of the dimension is the {\em indiscriminate} attacks, in which the attackers' goal is more flexible, such that misleading the classifier to any false positive categorization would be a success. 
		
		\item {\bf Security violation}
		This dimension distinguish the attacking strategy by the type of error that classifier is expected to make by it. At one end, the so-called {\em integrity} attack aims to cause false negatives on the classifier, which is similar to the goal of the {\em evasion attack}; At the other end, the so-called {\em availability} attack aims to would simply aims to cause the mis-classifications of both {\em false negatives} and {\em false positives}, so that the system becomes effectively unusable. 
		\end{itemize}
		
	These dimensions, along with their possible values, have considered many important points when working with adversarial attacks from the perspective of information security. In addition, more recent researches in deep learning community often consider another set of variables when discussing creating adversarial examples, which is well-summarized by Papernot et al. \cite{papernot2016limitations}. 
%	shown in Figure \ref{fig:threat-papernot}. 
%	\begin{figure}
%	\center
%	\includegraphics[width=0.65\textwidth]{threat-papernot.png}
%	\caption{The threat model taxonomy suggested by Papernot et al. \cite{papernot2016limitations}. }
%	\label{fig:threat-papernot}
%	\end{figure}
%	The vertical axis specifies the amount of knowledge that the attacker can obtain to craft the adverarials, and the horizontal axis specifies the complexity of the attacking goal. 
	This section will give more detailed discussions about some important aspects within this taxonomy. 
	
	\subsubsection{White-box and Black-box Attack}
	The difference between white-box and black-box attacks lies in whether the attacker has the knowledge of the classifier. In a white-box attack, the attacker would be able to have as mush knowledge as it need to make the adversarial perturbation, which is totally invisible black-box attack. 
	Such difference is partially captured by the {\em influence} dimension proposed by Barreno et al. \cite{barreno2006can}. 
	However, the {\em causative} attack specified by Barreno et al. launches the attack by influencing the classifier's behavior during the training process, while the white-box attack, though behave rather passively, would often require the gradient of the cost function with respect to some input $\boldsymbol{x}$. 
	
	% In between the black and white, some researches may impose some restrictions on the knowledge one may have about the classifier, leading to the {\em gray-box} attacks.
	The black-box attack, which often requires the attacker to explore the behaviors of the classifier, is similar to the {\em exploratory} attack specified by Barreno et al. On this side, different researches may have different assumption on the type of information that the probe input will return. For instance, in Papernot et al. setting, the probe will only returns the class label produced by the classifier \cite{papernot2016practical}, while in Narodytska and Kasiviswanathan setting, partial knowledge of the confidence score produced by the classifier can be obtained \cite{narodytska2017simple}. In this paper, we treat the White-box {\em v.s.} black-box as the most significant difference when presenting the selected strategy examples. 
	
	\subsubsection{Targeted and Un-targeted Attack}
	
	Unlike the interpretation of {\em targeted} attack in Barreno's taxonomy, in modern terms, however, a {\em targeted} attack commonly means that there exists a particular targeted class $t$ (the target class) which the classifier is expected to (mistakenly) output. This central confusion between the two settings is whether the attack {\em targets} on a particular input or output. Since in network security area, scenarios like intrusion detection is mostly binary classification problem, the targeted output is thus already implied. On the other hand, targeting at a specific output class becomes more valuable when tasks like image classification may involves thousands of possible classes. Another type of attack target is to move the correct class label outside the top-$k$ ranked output of the classifier as is proposed by Narodytska and Kasiviswanathan \cite{narodytska2017simple}. 
	One extreme example of the un-targeted attack task is proposed by Moosavi-Dezfooli et al. \cite{moosavi2017universal}, who show that it is possible to find a {\em single} small image perturbation to fool a deep neural network classifier on all natural images. 
	
	\subsubsection{Least-cost and Constrained Attack}
	
	Constructing the adversarial example requires a trade-off between the amount of perturbation to be added, and its utility (or effectiveness) of the attack that can be measured either by the likelihood of success or the desired impact on the probability distribution of the classifier's confidence. The $\ell^p$ distance defines a way to measure the perturbation between the original input and the distorted version, but does not address how the trade-off problem should be formalized. Broadly speaking, the attacker may either make sure the attack succeed when tries his best to minimize the amount of perturbation that attack requires on the input, or make sure the amount of perturbation is in a fixed range and tries his best to trick the classifier. Such difference leads to two broad types of attacking strategies, which are named as {\em best-effort} attack and {\em bounded} attack by this paper. 
	 
	%\paragraph{Least-cost attack}
	Least-cost attacks are of the type that would require the attack attempt {\em must} succeed, when minimize the amount of perturbation that is required. Since this formalization is used in the original paper on adversarial problem in deep learning by Szegedy et al. \cite{szegedy2013intriguing}, it is also the earliest and most widely used type of formalization. 
	They use the $\ell^2$ norm metric, the adversarial attack problem for a given input $\boldsymbol{x} \in \mathbb{R}^{m}$ on a target class $l$ can be described as Equation \ref{eq:l-bfgs}. 
	\begin{subequations}
	\begin{align}
		\text{Minimize~~} & \|\boldsymbol{r}\|_2 \text{ ,} \label{eq:l-bfgs-obj}\\
		\text{subject to:} & \nonumber \\
		& F(\boldsymbol{x} + \boldsymbol{r}) = l \text{, where } l \neq F(\boldsymbol{x}) \label{eq:l-bfgs-goal} \\
		& \boldsymbol{x} + \boldsymbol{r} \in [0,1]^m \label{eq:l-bfgs-bound}
	\end{align}
	\label{eq:l-bfgs}
	\end{subequations}
	Notice that the constrains expressed in Equation \ref{eq:l-bfgs-bound} is often known as the {\em box-constraint} to the optimization theory, and is always neccessary to make sure any pixel of the distorted input is still within the valid range of intensity, but sometimes omitted for simplicity. 
		
	%\paragraph{Bounded-cost attack}
	In some scenarios, the least-cost attack may not suffice the requirement of the attacker, who may impose more restrictions on the amount of perturbation allowed to be added to the input. The effectiveness of the perturbation is thus measured by how much confidence the classifier will place on the distorted version comparing to the original input, when the maximum allowable distortion in the optimal direction is added. Tasks of this type can be formalized as 
	\begin{subequations}
	\begin{align}
		\text{Maximize~~} &  f_l (\boldsymbol{x} + \boldsymbol{r}) \text{ ,} \label{eq:bounded-obj} \\
		\text{subject to:} & \nonumber \\
		& \| \boldsymbol{r} \| \le L \label{eq:bounded-constraint} \\
		& \boldsymbol{x} + \boldsymbol{r} \in [0,1]^m \text{ ,}
	\end{align}
	\label{eq:bounded-attack}
	\end{subequations}
in which $f_l(\cdot)$ denotes the probability score the neural network assigns to the class label $l$ for any given input. The hyper-parameter $L$ specifies the upper-bound of the perturbation that can be added to any input instance $\boldsymbol{x}$. 
	
	Notice that when comparing to Equation \ref{eq:l-bfgs-obj}, the optimization objective in Equation \ref{eq:bounded-obj} becomes the attack utility rather than the perturbation, while the amount of perturbation now appears as a constraint term in Equation \ref{eq:bounded-constraint}. 
	Within the attacking bound $L$, one may still attempt to find the optimized solution $\boldsymbol{r}^{*}$ for all feasible points of $\boldsymbol{r}$. For attacking strategies designed for least-cost attacks that uses iterative update method, it is simple to employ {\em early-stop} mechanism such that the iteration will end whenever the amount of distortion is above the bound. 
	
%	\subsubsection{Online or offline}
%	Online learning allows the learner to adapt to changing conditions; the assumption of stationarity is weakened to accommodate long-term changes in the distribution of data seen by the learner. Online learning is more flexible, but potentially simplifies causative attacks. 
	
	\subsection{White-box Attacks}
	
	White-box attack assumes that the attacker has the complete knowledge of the classifier. In most of the strategies, the attack strategy would need to be able to find the derivative (or gradient) of the model at certain input point. Although such conditions may seem impractical in most real-world scenarios, this type of attacks does reveal some underly weakness of the classifiers, and also forms the basis for the more realistic black-box attacks. 
	
	In this part of the paper, we introduce several representative adversarial attack strategies that have been proposed by previous researches. Since it is unlikely to enumerate all possible strategies that have been seen in the literature, this survey tries to capture the basic ideas in the most cited works within the deep learning community, and their relationships.

	\subsubsection{The Optimization Attacks}
	
	The L-BFGS attack is the firstly proposed by Szegedy et al. along with their discovery of the adversarial examples for the deep learning algorithms \cite{szegedy2013intriguing}. They formalize the problem as finding the optimal perturbation input $\boldsymbol{r}$, in terms of the $\ell^2$ norm, to make the classification function output a wrong class, which can be described in Equation \ref{eq:l-bfgs}.	
	They replace the attack goal $F(\boldsymbol{x} + \boldsymbol{r}) = l \text{~where~} l \neq F(\boldsymbol{x})$ by introducing the loss function that originally used to optimize the parameters $\mathrm{loss}_{f,l}(\boldsymbol{x})$, and tackle this problem by using the box-constrained L-BFGS, and rewrite the optimization problem into the form of
	\begin{equation}
	\begin{split}
		\text{Minimize~~} & c \cdot \| \boldsymbol{r} \|^2_2 + \mathrm{loss}_{f,l}(\boldsymbol{x} + \boldsymbol{r}) \text{ ,} \\ 
		\text{such that: } & \\
		& \boldsymbol{x} + \boldsymbol{r} \in [0, 1]^m \text{ .}
	\end{split}
	\end{equation}
	By using a line search to find the constant $c$, this method yields an approximation to the optimal solution due to the non-convexity nature of deep neural networks. 
	
	The optimization-based attack strategy proposed by Carlini and Wagner \cite{carlini2017towards}, referred to as the {\em CW} attack, reuse the task definition as is described in Equation \ref{eq:l-bfgs}, and can be regarded as an extension of the L-BFGS attack \cite{szegedy2013intriguing} in several ways. 
	First, in addition to the $\ell^2$ norm metric, they define a more general distance function $\mathcal{D}(\boldsymbol{x}, \boldsymbol{x}')$, which can has any of the $\ell^0$, $\ell^2$, or $\ell^{\inf}$ form. 
	Second, instead of using the loss function of the classifier $\mathrm{loss}_{f}$ in substitution of the target directly, they study up to seven different objectives. 
	Third, instead of using the L-BFGS-B optimization algorithm, they investigate three different methods for the optimization problem that
	% grammar error here? 
	make using optimizers don't natively support box constraints possible.
	
	\subsubsection{The Fast Gradient Sign Method}
	
	One important drawback of the original L-BFGS method is its huge computation cost, which requires estimating the Hessian matrix. Therefore, many follow-up researches start to use only the first derivative information, and lead to a wide class of the {\em first-order method}. Fast Gradient Sign Attack (FGSM) \cite{goodfellow2014explaining} is among the first a few proposed adversarial attacks against deep learning algorithms, which aims to optimize the $\ell^1$ distance metric, and is designed primarily to be {\em fast} by using only the gradient information, instead of producing very close adversarial examples.
	 
	Given the input $\boldsymbol{x}$, the fast gradient sign method sets the adversarial example as
	\begin{equation}
		\boldsymbol{x}' = \boldsymbol{x} + \epsilon \cdot \mathrm{sgn} \left( \nabla_{\boldsymbol{x}} \mathrm{loss}_{f, y_\text{true}}(\boldsymbol{x}) \right) \text{ ,}
	\end{equation}
	where $y$ is the true label for the input $\boldsymbol{x}$, $\epsilon$ is chosen to be sufficiently small so as to be undetectable. Intuitively, the distorted input $\boldsymbol{x}'$ is obtained by {\em moving away} from the original input point of $\boldsymbol{x}$ to increase the loss function alongside the direction of the gradient. This method can be easily used to generate {\em targeted attack} by making the distorted input {\em move towards} the target class at the gradient's direction, as is described in Equation \ref{eq:fgsm-targeted}, where $y'$ is the targeted output class label. 
	\begin{equation}
		\boldsymbol{x}' = \boldsymbol{x} - \epsilon \cdot \mathrm{sgn} \left( \nabla_{\boldsymbol{x}} \mathrm{loss}_{f, y'}(\boldsymbol{x}) \right) \text{ .}
		\label{eq:fgsm-targeted}
	\end{equation} 
	
	Tram{\`e}r et al. extend this attacking strategy by adding a random noise that follows Gaussian distribution before the standard FGSM \cite{2017arXiv170507204T}.
	Kurakin et al. \cite{kurakin2016adversarial_physical} extend the FGSM by replacing the single update with step size $\epsilon$ in the direction of $\nabla_{x} \mathrm{loss}$ with multiple updates of step size $\alpha$ and clipped by the same $\epsilon$. Their proposed basic iterative method can be expressed as
	\begin{equation}
		\boldsymbol{x}'_{i} = \mathrm{Clip}_{\boldsymbol{x}, \epsilon} \{\boldsymbol{x}'_{i-1} + \alpha \cdot \mathrm{sgn} (\nabla_{\boldsymbol{x}} \mathrm{loss}_{f, y} (\boldsymbol{x}_i)) \}
	\end{equation}
	where $\alpha = 1$, i.e. the value of each pixel changes only by 1 on each step.  
	They then further propose an iterative {\em least-likely class} method, which turns the original {\em indiscriminate} attack strategy to an {\em targeted} strategy. The least-likely class $y_{LL}$ according to the predication of the trained network on image $\boldsymbol{x}$ is
	\begin{equation}
		y_{LL} = \arg\min_{y} p(y | \boldsymbol{x}) \text{ .}
	\end{equation}
	
	Analogous to the change in the one-step version from un-targeted FGSM to targeted FGSM, the iterative version of targeted FGSM can be described as:
	\begin{equation}
		\boldsymbol{x}'_i = \mathrm{Clip}_{\boldsymbol{x}, \epsilon} \{\boldsymbol{x}'_{i-1} - \alpha \cdot \mathrm{sgn} (\nabla_{\boldsymbol{x}} \mathrm{loss}_{j, y_{LL}}(\boldsymbol{x})) \}
	\end{equation}
	
	\subsubsection{The JSMA Attack}
	Another type of first-order method called Jacobian-based Saliency Map Attack (JSMA) is proposed by Papernot et al. \cite{papernot2016limitations}. 	
	The idea of {\em saliency map} originates from the visualization method for convolutional networks proposed by Simonyan et al. \cite{DBLP:journals/corr/SimonyanVZ13}. 
	In JSMA, the saliency map identifies most promising input feature on which perturbation should be applied to cause the desired changes in the classifier's output. Adversarial examples are then generated by performing greedily feature-wise modification on the  input instance. 
	
	The saliency map is based on the {\em forward derivative}, which refers to the gradient of the forward-pass function $f$ with respect to the input $\boldsymbol{x}$. 
	The ``positive'' salience score of the feature $i$ is given by
	\begin{equation}
	S(\boldsymbol{x}, t)[i] = \left\{
	\begin{array}{ll}
		0 & \begin{array}{l}
		\text{ if } \frac{\partial f_t(\boldsymbol{x})}{\partial \boldsymbol{x}_i} < 0 \text{ , or } \\ \sum_{j \neq t} \frac{\partial f_j(\boldsymbol{x})}{\partial \boldsymbol{x}} > 0 \end{array}
		\\
		\frac{\partial f_t(\boldsymbol{x})} {\partial \boldsymbol{x}} \left| \sum_{j \neq t} \frac{\partial f_j(\boldsymbol{x})}{\partial \boldsymbol{x}_i} \right| & \text{otherwise.}
	\end{array}
	\right.
	\end{equation}
	
	That is, the saliency score of feature $i$ remains 0, if increasing its value will either decrease the confidence of $\boldsymbol{x}$ being classified as $t$, or increasing its value will cause, averagely, the increase of the confidence score of all other output classes that $j \neq t$. Otherwise, its saliency score is the forward derivative 
	$$\frac{\partial f_t(\boldsymbol{x})}{\partial \boldsymbol{x_i}}$$
	scaled by the factor of 
	$$\left| \sum_{j \neq t} \frac{\partial f_j{\boldsymbol{x}}}{\partial \boldsymbol{x}_i} \right| \text{.}$$ It is also easy to create a negative version of the saliency map by considering the impact of decrease the value of feature $i$:
	\begin{equation}
	S(\boldsymbol{x}, t)[i] = \left\{
	\begin{array}{ll}
		0 & \begin{array}{l}
		\text{if } \frac{\partial f_t(x)}{\partial x_i} > 0 \text{ , or } \\ \sum_{j \neq t} \frac{\partial f_j(x)}{\partial x} < 0 \end{array}
		\\
		( \frac{\partial f_t(x)} {\partial x} ) \left| \sum_{j \neq t} \frac{\partial f_j(x)}{\partial x_i} \right| & \text{otherwise. }
	\end{array}
	\right.
	\end{equation}
	The overall adversarial generation algorithm is implemented by iteratively choose the current-optimal feature to modify until either the attack succeed or meet the distortion bound $\Upsilon$. 
	
	\subsubsection{The DeepFool Attack}
	The DeepFool attacking strategy proposed by Moosavi-Dezfooli et al. \cite{moosavi2016deepfool} has a lot in common with the iterative FGSM, in the sense that it also moves towards the gradient's directions iteratively, and updates the gradient at each iteration. 
	However, the Deepfool strategy uses the original gradient direction instead of the sign of gradient. In addition, instead of setting the change rate to a fixed value $\alpha$, which is set to 1 at iterative FGSM's experiment\cite{kurakin2016adversarial_physical}, the Deepfool strategy will always move the attacking point to the estimated decision boundary directly, by following the current gradient's direction with the step size of the estimated distance between its current position and the decision boundary. 
	So the Deepfool strategy for a binary classifier in the $\ell^2$ criterion setting can be described in Equation \ref{eq:deepfool-uni}:
	\begin{equation}
	\begin{split}
		&\boldsymbol{r}_i = -\frac{f(\boldsymbol{x}_i)}{\| \nabla f(\boldsymbol{x}_i) \|_2^2} \nabla f(\boldsymbol{x}_i) \\
		&\boldsymbol{x}_{i + 1} = \boldsymbol{x}_i + \boldsymbol{r}_i \\
		&\hat{\boldsymbol{r}} = \sum_i{\boldsymbol{r}_i}
	\end{split}
	\label{eq:deepfool-uni}
	\end{equation}
	where $\boldsymbol{x}_i$ denotes the distorted input at iteration $i$, $\frac{f(\boldsymbol{x}_i)}{\| \nabla f(\boldsymbol{x}_i) \|_2}$ is the estimated distance between the point of $\boldsymbol{x}_i$ and the decision boundary of the function $f$, while $- \frac{\nabla f(\boldsymbol{x}_i)}{\| \nabla f(\boldsymbol{x}_i) \|_2}$ is the normalized gradient vector pointing towards the decision boundary. The same method can be easily extended to a multi-class classifier. 
	
%	described in Equation \ref{eq:deepfool-multi}, which is similar to a sequential convex programming where the constraints are linearized at each step.
%	\begin{equation}
%	\begin{split}
%		& w'^{k}_i = \nabla f^k(x_i) - \nabla f^{\hat{k}(x_0)}(x_i) \\
%		& f'^{k}_i = f^k(x_i) - f^{\hat{k}(x_0)}(x_i) \\
%		& \hat{l}_i = \arg\min_{k \neq \hat{k}(x_0)} \frac{|f'^{k}_i|}{\|w'^k_i\|_2} \\
%		& r_i = \frac{f'^{\hat{l}}_i}{\| w'^{\hat{l}}_i \|} w'^{\hat{l}}_i \\
%		& x_{i+1} = x_{i} + r_{i} \\
%		& \hat{r} = \sum_i {r_i}
%	\end{split}
%	\label{eq:deepfool-multi}
%	\end{equation}

%	\subsubsection{The ``universal'' attack}
%	
%	\cite{moosavi2017universal} show the existence of a universal (image-agnostic) and very small perturbation vector that causes natural images to be misclassified with high probability.
%	
%	The main focus of this paper is to seek perturbation vectors $v \in R^{d}$ that fool the classifier $\hat{k}$ on almost all data-points sampled from $\mu$. That is, we seek a vector $v$ such that
%	
%	\begin{equation}
%		\hat{k}(x + v) \neq \hat{k}(x) \text{  for ``most''  } x \sim \mu.
%	\end{equation}
%	
%	We coin such a perturbation universal, as it represents a fixed image-agnostic perturbation that causes label change for most images sampled from the data distribution $\mu$.
%	
%	universal perturbations have a remarkable generalization property, and also generalize well across deep neural networks. Such perturbations are therefore doubly universal, both with respect to the data and the network architectures.
	
	\subsection{Black-box Attacks}
	\label{sec:black-box}
	
		\subsubsection{Intuitions}
		
		Crafting adversarial attack does not make random moves, but requires heuristics derived from the knowledge of the targeted classifier. Therefore, the main problem for making the black-box attacks is how to obtain the information needed. The two basic premises making it possible are the transferability of the adversarial examples and probing the behavior of the classifier. 
		
		\paragraph{Transferability}
		Early researches on the adversarial problem in deep learning has shown that an adversarial example for one model will often transfer to be an adversarial on a different model, even if they are trained on different sets of training data \cite{szegedy2013intriguing, goodfellow2014explaining}. 
		A more recent research by Papernot et al. \cite{papernot2016transferability} shows a more astonishing property of the adversarial examples for the machine learning algorithms, that they can transfer between the models even if they use entirely different algorithms (i.e., adversarial examples on neural networks transfer to random forests). They argue that for each pair of classifiers $ (F, F') $: the transferability can be broken down into two types: the {\em intra-technique transferability}, meaning both $F$ and $F'$ are of the same machine learning technique (e.g., both are neural networks) but are trained on different parameter initializations or datasets, and the {\em cross-technique transferability}, meaning $F$ and $F'$ are two different techniques in natural (e.g., $F$ is based on a neural network and $F'$ a decision tree). Then they prove that both types of the transferabilities are consistently strong across the space of machine learning techniques. 
		
		\paragraph{Probing}
		
		Another important premise of the black-box attack is that the attacker is usually assumed to be able to have probe the classifier. To avoid ambiguity, we would refer to the targeted classifier as the {\em oracle}. A model receives $m$ dimensional input, each with $k$ possible values has the input space of cardinality of $m^k$. Exploring the behavior of the oracle by testing all possible input is intractable. Therefore, approaches that require probing commonly focuses how to find the minimal set of input points to optimally probe the oracle classifier. More details on the probing strategies will be discussed in Section \ref{sec:substitute-network}. 
			
		\subsubsection{Substitute Network}
		\label{sec:substitute-network}
		
		Based on the transferability of the adversarial examples, Papernot et al. proposed a substitute network approach \cite{papernot2016practical} which trains a local substitute deep neural networks with only limited number of label queries and can be scaled to different machine learning classifier types. 
		
\begin{figure}[h]
			\center
    		\includegraphics[width=0.45\textwidth]{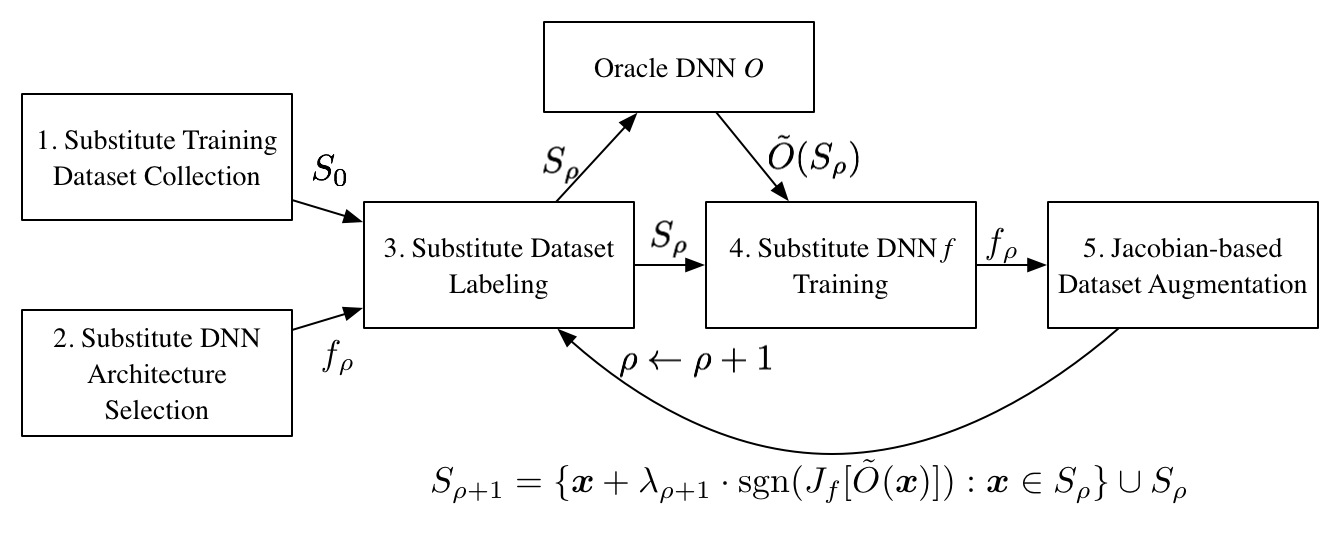}
    		\caption{The workflow of the substitute network as is proposed by Papernot et al.\cite{papernot2016practical}. }
    		\label{fig:substitute-network}
		\end{figure}

		For an oracle network $O$ that outputs the probability distribution to indicate the final class label $\tilde{O}(\boldsymbol{x}) = \argmax_{j} O_j(\boldsymbol{x})$, the attack process begins with the construction of a substitute network $f$, which should comply with the oracle's expected input and output.\footnote{As before, the class label output of $f$ is given by $F(\boldsymbol{x}) = \argmax_j f_j(\boldsymbol{x})$.} To make the substitute network best imitate the behavior of the oracle with the only limited label queries, they generate synthetic training data from an initial set of seed points, based on the Jacobian matrix $\mathbf{J}_f$ to identify the directions in which the model’s output varies most significantly. For each of the input point $\boldsymbol{x}$, a new instance can be generated as $\lambda \cdot \mathrm{sgn}(\mathbf{J}_f(\boldsymbol{x}) [\tilde{O}(\boldsymbol{x})])$. 
		The overall working process is illustrated in Figure \ref{fig:substitute-network}, in which $S_0$ is the initial seed training data, $f$ is the created substitute network architecture. 		
		
		At iteration $\rho$, the algorithm query the oracle $O$ to obtain the output label $\tilde{O}(S_\rho)$, together with the inputs $S_\rho$, to train a new version of the substitute network $f_\rho$. At the end of the iteration $\rho$, a new set of training instances can be synthesized by appending distorted version of each instance $\boldsymbol{x} \in S_\rho$ to $S_{\rho+1}$. 
		To prevent the exponential growth of the query candidates, the reservoir sampling technique can be adopted here \cite{papernot2016practical, papernot2016transferability}. For some iteration $\rho$ that above a threshold $\sigma$, only $\kappa$ number of new inputs are selected from the original Jacobian-based dataset augmentation, such that each input in $S_{\rho-1}$ has an equal probability $\frac{1}{S_{\rho-1}}$ to be augmented in $S_{\rho}$.

%		\subsubsection{Ensemble approach}
%		\cite{liu2016delving} propose novel ensemble-based approaches to generating transfer- able adversarial examples.
%		approximate the minimal distortion $B$ along a direction $\delta$, such that $x B = x + B \delta$ generated for $M_1$ is adversarial for both $M_1$ and $M_2$. Here $\delta$ is the direction, i.e., $sgn(\nabla_{x}\ell)$ for FGS, and $\nabla_{x}\ell / \| \nabla_{x} \ell \|$ for FG.
%		adversarial image remains adversarial for multiple models, then it is more likely to transfer to other models as well.
		
		\subsubsection{Local Search}
		
		Narodytska and Kasiviswanathan \cite{narodytska2017simple} point out that the substitute network method \cite{papernot2016practical} tend to result in a degradation in the effectiveness of the attacks. Meanwhile, their proposed {\em local search} method does not require the assumption of the transferability, and can also save the overhead of retraining the substitute network. However, their local search method would require partial information of the confidence score produced by the neural network, when the substitute network method by Papernot et al. needs only the class label output of the oracle classifier. 
		
		They target on a so-called $k$-misclassification task. If we reuse the notations in Section \label{sec:substitute-network}, that $O(\boldsymbol{x})$ returns the confidence score for each of the class by the oracle, and $\tilde{O} = \argmax_j O_j(\boldsymbol{x})$ returns the final output label of the oracle, then $\pi(O(\boldsymbol{x}), k)$ is a function that returns confidence ranking of up-to top-$k$ classes in a vector form, which is also called the $\pi$ vector. Then the $k$-misclassification can be described as 
		\begin{equation}
			\tilde{O}(\boldsymbol{x}) \notin \pi(O(\boldsymbol{x}), k) \text{ .} 
		\end{equation}	
		
		In general, the proposed local search strategy relies on the numerical approximations of the gradient, and iteratively searches a local neighborhood of pixels positions, to refine the current image. 
		Each iteration of the local search procedure consists of two steps: first, evaluate the objective function $f_\text{obj}$ for each of the points $Z = \{ \hat{\boldsymbol{z}}_1, \dots, \hat{\boldsymbol{z}}_n \}$ that form a local neighborhood, which is defined as the confidence score of the original correct class label of the on the distorted version of the input that produced by the oracle; and then, select a new solution $\boldsymbol{z}_i$ based on the previous solution $\boldsymbol{z}_{i-1}$ and points in $Z$. 
		% For the input image $\boldsymbol{x}$, they use a simple objective function defined as $f_{\tilde{O}(\boldsymbol{x})}(\boldsymbol{x}') = \max_j O_j(\boldsymbol{x})$. 
		
		\subsubsection{GAN-based Attack}
		\label{sec:gan-attack}
		The $\ell^p$ interpretation of the ``imperceptible'' provides a mathematically convenient way to measure the distortion to be added to the input. However, the simple $\ell^p$ norm measures only the distortion in the superficial high-dimensional space, which may not reflect the impact on the human's perception level, which can be crucial in complex domains. For instance, simple errors in the gramma or the improper wording in the adversarial example in natural language can be outstanding, as has been noticed by Li et al. \cite{li2016understanding} and Jia et al. \cite{jia2017adversarial}. Zhao et al. considers creating the natural adversarial example as their major focus, which can be tackled by using the GAN model \cite{zhao2017generating}. One nice feature of their approach is that it requires no information of the gradient of the classifier, and thus can be applied in a black-box scenario. 
		
		As has been introduced in Section \ref{sec:gan}, generative adversarial networks is family of generative models that are capable of generating realistic samples of various types of input like image or text from a latent space input $\boldsymbol{z}$. Zhao et al. first train a generator $\mathcal{G}_\theta$ parameterized by $\theta$ by using a W-GAN on the training dataset, which maps a random dense vector $\boldsymbol{z} \in \mathbb{R}^d$ to an instance in the superficial space $\boldsymbol{x} \in \mathcal{X}$. Then they separately train a inverse function $\mathcal{I}_{\gamma}$ parameterized by $\gamma$ which is capable of mapping the superficial space to the latent space by minimizing the reconstruction error:
		\begin{equation}
			\min_{\gamma} \mathbb{E}_{\boldsymbol{x} \sim \mathbb{P}_\text{real}(\boldsymbol{x})} \| \mathcal{G}_{\theta}(\mathcal{I}_{\gamma}(\boldsymbol{x})) - \boldsymbol{x} \| + \lambda \mathbb{E}_{\boldsymbol{z} \sim p_{\boldsymbol{z}}(\boldsymbol{z})} \| \mathcal{I}_{\gamma}(\mathcal{G}_{\theta}(\boldsymbol{z})) - \boldsymbol{z} \| \text{ .}
		\end{equation}
		Based on the learnt generator $\mathcal{G}_\theta$ and its inverse function $\mathcal{I}_\gamma$, the original input $\boldsymbol{x}$ can be mapped into the latent space $\boldsymbol{z} = \mathcal{I}_{\gamma}(\boldsymbol{x})$, on which the perturbation is performed, resulting the distorted version in the latent space ${\tilde{\boldsymbol{z}}}$, and then mapped back to the input space $\tilde{\boldsymbol{x}} = \mathcal{G}_{\theta}(\tilde{\boldsymbol{z}})$. Thus, the natural adversarial example $\boldsymbol{x}^*$ is derived by:
		\begin{equation}
		\begin{split}
			& \boldsymbol{x}^* = \mathcal{G}_{\theta}(\boldsymbol{z}^*) \\
			& \text{where~~} \boldsymbol{z}^* = \argmin_{\tilde{z}} \| \tilde{z} - \mathcal{I}_{\gamma}(\boldsymbol{x}) \| \text{~~s.t.~~} F(\mathcal{G}_{\theta}(\tilde{z})) \neq F(\boldsymbol{x}) \text{ ,}
		\end{split}
		\end{equation}
		and $F$ denotes the original classifier the attacker is targeting at. Although the ``naturalness'' is hard to measure directly, sample cases produced from several standard dataset do suggest their works as expected on both image and text domain. 
		
\begin{figure}[h]
		\center
		\includegraphics[width=0.45\textwidth]{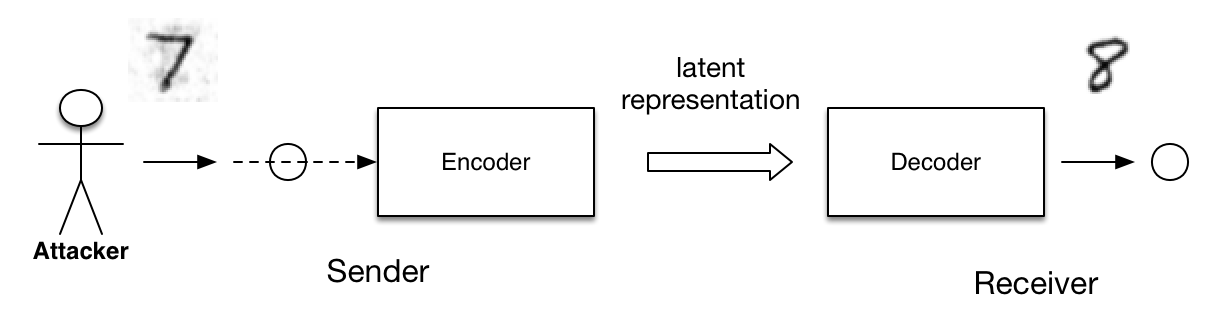}
		\caption{An illustration of the adversarial attack scenario of \cite{kos2017adversarial}. The encoder compressed the original input into a latent representation $\boldsymbol{z}$, which is then de-compressed by a decoder. }
		\label{fig:adv-vae}
		\end{figure}

		A similar idea to this GAN based adversarial generation approach is proposed by \cite{kos2017adversarial}, in which a deep generative model is involved, and the attacker aims to manipulate the latent representation $\boldsymbol{z}$ instead of the superficial input $\boldsymbol{x}$. However, their attacking scenario of \cite{kos2017adversarial} is quite different, in that instead of attacking a single classifier to produce a wrong label, the adversarial input is constructed for an encoder-decoder pair, such that the decoding output becomes dramatically different from the input, as illustrated in Figure \ref{fig:adv-vae}.

%%%%%%%%%%%%%%%%%%%%%%
%
% Beginning of Chapter IV
% 
%%%%%%%%%%%%%%%%%%%%%%
		
\section{Defending Strategies}
	
	\subsection{Intuitions}
	
	As their counter part of the attacking strategy taxonomy, Barreno et al. also propose a categorization for the defense strategies \cite{barreno2006can}, which shares much overlapping with the ideas of the latest defending method for contemporary deep learning algorithms:
	\begin{itemize}
		\item {\bf Robustness} Whereas the entire problem of adversarial training can be boiled down to the robustness, Barreno et al. uses this term here to refer to a more specific meaning of {\em model regularization}. Regularization extends the basic optimization objective by adding a term to penalize the model complex, and has now become a standard practice in machine learning.
		\item {\bf Detecting Attacks} The defender may devise an attack detector to filter out the abnormal input before sending to the classifier. 
		\item {\bf Disinformation} The defender may try to confuse the adversary’s estimate of the learner’s state. For instance, a sophisticated learner could trick the adversary into believing that a particular intrusion was not included in the training set.
		\item {\bf Randomization} The defender can randomize the decision boundary of the classifier, which is especially effective to defend against targeted attack that focuses on a specific input instance.  
	\end{itemize}
	
	More recent researches on adversarial in deep learning community have proposed various methods to enhance the robustness of the model, most of which can be classified into one or more approach categories listed above. 
	Defending strategies, once published, are exposed to the public, and can be compromised at any moment. In fact, many of the strategies to be reviewed in this section have already been proved to be unsecured by some follow up studies, and even the existence of universally robust defending strategy is by far an open question.  
	Thus, some researchers try to provide intuitive interpretations to the cause of adversarial inputs, which may hopefully lead to some theoretical guides to the design of new and more effective defending strategies. 
	
	\subsubsection{The Capacity Matters}
	
	When adversarial were firstly spotted in the deep neural networks, Szegedy et al. suggest that such phenomenon was caused by the extreme non-linearity of the function. However, follow-up researches prove that compared to the deep learning models, simple models with low model capacity are in fact more vulnerable to such attacks \cite{goodfellow2014explaining, kurakin2016adversarial_scale}.
	This argument is also soon supported by Madry et al. \cite{madry2017towards}, who show that the model capacity alone is a major impactor on improving the robustness of the classifier, even training on the natural dataset only. In a word, ``capacity alone helps''. 
	
	Theoretically speaking, the {\em universal approximator theorem} \cite{hornik1989multilayer} guarantees that a neural network with at least one hidden layer can represent any function to an arbitrary degree of accuracy, provided that hidden layer has enough units. This gives the deep neural networks the possibility to be able to resist the adversarial perturbation given proper training data and procedure. On the other hand, it is impossible to have a shallow linear model to keep constant near training points while also assigning different outputs to different training points, making the models of this kind to have the capability to resist adversarial input rather dim. Therefore, it has been so far widely accepted that, shallow models are however, even more vulnerable to adversarial attacks. The concerns of the risk in deep models may partly due to its complexity --- it may hard to fix a known problem even if it has been spotted.

	\subsubsection{Flaws in the Training Procedure}
	
	Although the root cause of the existence of adversarial examples for deep learning models is still unclear, it has been widely accepted that it has some direct connection with the standard supervised training procedure that does not consider the potential adversarial attacks.  
	%If we study the optimization objective for deep learning models carefully, it wouldn't be difficult to see t
	The objective defined by deep learning algorithms is usually set to minimize the expected error (or cost) that the classifier would output. This objective is sensible only under certain assumptions:
	\begin{enumerate}
		\item The training objective is usually the empirical risk minimization (ERM) that averages over all known training instances, in which each of instance is considered equally important. Though one may specify the weight of each training instance or the output labels, it is still a problem how these weights can be derived. 
		\item More importantly, the training data is well-sampled in a way that can unbiasedly reflect the data distribution which is assumed to be stationary. This does not reflect the dynamics of the mini-max game that the attacker and defender should continuously adjust their strategies. 
	\end{enumerate}
	In the scenarios when such assumptions no longer hold, extreme caution must be paid to a better-formulated objective. Previous researches on alternative views of the optimization objectives usually root in the need to balance the {ability to generalize} from the training data, which is considered to be insufficient to problem of the adversarial learning \cite{goodfellow2014explaining}. 

	% For instance, one may specify the weight of each training instance or the output labels. However, the follow-up questions arises that how do we decide these weights as hyper-parameters? Another way to take into account the imbalance of the data or label is to use the measurements like Precision/Recall or Type-I/II error. But these measurements are used in test-time only, and is not the optimization objective during the training process. 

	\subsection{Strengthening the Training Process}
	
	As is discussed above, one paradigm to understand the existence of adversarial attack is due to the indiscretion within the {\em training process}. Consequently an intuitive way to figuring out the solution is to improving the training process somehow to address the potential adversarial inputs. 
	
		\subsubsection{Data Augment}
		
    	Szegedy et al. \cite{szegedy2013intriguing} showed that by training on a mixture of adversarial and clean examples, a neural network could be regularized somewhat. 
		Training on adversarial examples is different from other data augmentation schemes. For example, common approach to augment data for training a computer vision like translations or crops are expected to actually occur in the test set. 
		However, for adversarial training, data augmentation requires inputs that tend to expose flaws in the ways that the model conceptualizes its decision function, which are unlikely to occur naturally. 

    	Goodfellow et al. recommend that the training process needs to continually generating new adversarial examples at every step, and inject these adversarial examples into the training set \cite{goodfellow2014explaining}. 
		% Adversarial training was originally developed for small models that did not use batch normalization. 
		They also suggest to use batch normalization \cite{ioffe2015batch} to scale adversarial training to large models for tasks like ImageNet, 
		in which it is critical for each batch of examples to contain both normal and adversarial examples. 
		
		The effectiveness of the data augment method is largely decided by the way that the augmented data are generated. 
		Madry et al. \cite{madry2017towards} argue that adversarial training may benefit the most, when the adversarial examples for training closely maximize the model's loss. They advocate to use the multi-step PGD (projected gradient descent) as the best first-order method to find the perturbation value to generate these adversarial examples, as well as to increase the model capacity simultaneously. This approach has been considered one of the most effective defending strategies that has been developed by far. 

    	Data augment is a flexible method that can be easily plugged into other defending mechanisms. 
		For instance, it is possible to inject the auto-constructed adversarial examples into the training data dynamically and iteratively, while employing other counter-adversarial methods such as the distillation or test-time detection techniques.

		\subsubsection{Extra-penalty Term}
		The simplest form of regularization for neural networks, commonly known as the {\em weight-decay} is by penalizing the weights between layer pairs. Goodfellow et al. show that the original weight decay regularizer can over-estimate the impact of adversary examples when compared to the data-augment approach, making it hard to balance the training error rate and the regularization benefit. \cite{goodfellow2014explaining} As such, they suggest an alternative way to define the adversarial learning objective, based on the fast gradient sign method, which can be expressed by Equation \ref{eq:fgsm-defend}:
		\begin{equation}
			\tilde{\mathrm{loss}}(\theta, x, y) = \alpha \cdot \mathrm{loss}(\theta, x, y) + (1 - \alpha) \cdot \mathrm{loss}(\theta, x + \epsilon \cdot sign( \mathrm{loss} )) \text{ ,}
			\label{eq:fgsm-defend}
		\end{equation}
		in which the hyper-parameter $\alpha$ is set to 0.5 by default. This approach can be interpreted in a way that it continually injects the newly generated adversarial examples based on the current version of the model.
		
		\subsubsection{Distillation}
		
		The transferability of the knowledge between models may not only exist for adversarial examples, but also allow us to compress models \cite{Bucilua:2006:MC:1150402.1150464}. The idea of distillation is first proposed by Hinton et al. \cite{hinton2015distilling}, in order to ``distill'' the knowledge from a {\em cumbersome} model to a small model that is more suitable for deployment. 
		
		For a classification task, the class probabilities produced by the neural networks are typically from a $softmax$ layer that converts the logit, $z_i$ into a normalized probability $q_i$:
		\begin{equation}
			q_i = \frac{e^{{z_i}/{T}}} {\sum_j e^{{z_j}/{T}}} \text{ ,}
		\end{equation}
		where the temperature parameter $T$ is commonly set to 1 by default, and a higher value for $T$ always produces a softer probability distribution over classes.  
		The ground-truth labels in the training data commonly follow a distribution $p$ that puts all mass one a single correct value (i.e., the Dirac distribution of Equation \ref{eqn:dirac}). The training loss is commonly defined by the cross entropy between $p$ and the $softmax$ probability $q$, which is sometimes improper (e.g., a digit ``7'' may sometimes does look like ``1'') and may also cause numerical issues. Therefore, the distilling training process trains a cumbersome model (also known as the {\em teacher model}) first as usual, which outputs a {\em softer version} of the target distribution to train a distill model. Both the training process will use a higher temperature value, and restore the temperature to 1 for the test cases. 
			
		The intuition behind using distill training as a defense strategy by Papernot et al. \cite{papernot2016distillation} is to overcome the possible over-fitting which may cause the neural networks to have ``blind spots'' \cite{szegedy2013intriguing}. Comparing to the original distill training, defensive distillation employs the same network architecture for the distill model as the teacher model, meanwhile use a large temperature parameter $T$ during training. 
				
	\subsection{Test Time Perturbation Recovery}
	\label{sec:pert-recovery}
	
	Another class of approaches tries to tackle the problem {\em at test time} by removing the adversarial distortion from the input, usually follow the general idea to map the input into a latent space, and then reproduce a similar input via a deep generative model. In this part of the paper, we will cover several representative works that follow this intuition. 
		
		\subsubsection{Deep Contractive Network}
		
		Comparatively speaking, signal de-noising is a well-studied research topic in machine learning area. Suppose we regard the adversarial perturbation on the input image as a special type of ``noise'', then is it possible to remove such noise by standard de-noising techniques as a pre-processing stage before classification? 
		
		This question is firstly answered by Gu and Rigazio \cite{gu2014towards}, who show that a Deep Contractive Network (DCN) is a suitable choice for this task. As baseline approaches, they consider three simple methods for comparison:
		\begin{enumerate}
		\item by adding additive Gaussian noise and Gaussian blurring to the input;
		\item by constructing a standalone auto-encoder, in order to map the adversarial examples back to the original data samples; and 
		\item by using a standard denoising autoencoder (DAE), which can be regarded as the combination of the previous two methods 
		\end{enumerate}
		
		Their empirical experiments show that, the first baseline method can recover a large portion of the adversarial examples, but still falls drastically behind the performance on the clean input. 
		In contrast, the second method can recover at least 90\% of the original adversarial errors, but new adversarial examples can again generated from the stacked network that consists of both the encoding neural network and the original classifier. 
		The same issue is also shared by the third method, making these methods not appropriate choice for the task. In fact, the newly derived models can be even more vulnerable to adversarial attacks. 
		
		As such, they appeals for an alternative type type of autoencoder, namely the Contractive autoencoder (CAE), which employs an extra Frobenius norm of the Jacobian matrix $\lambda \cdot \| \mathbf{J} \|_2 = \lambda \cdot \| \nabla_{\boldsymbol{x}} \boldsymbol{z} \|_2$
%		\begin{equation}
%			\lambda \cdot \left\| \frac {\partial \boldsymbol{z}} {\partial \boldsymbol{x}} \right\|_2 \text{, }
%		\end{equation}
		in addition to the reconstruction error appears in the loss function of autoencoder \cite{Rifai2011Contractive}, where $\boldsymbol{z}$ is the code in latent space produced by the encoder, and $\lambda$ is the trade-off hyper-parameter. With this penalty term, a contractive autoencoder tries to minimize the reconstruction error while keep the model {\em invariant} to any small change on the input. 
		Deep Contractive Network (DCN) generalizes the (shallow) Contractive Autoencoder (CAE) by using a deep feed-forward neural network for both the encoder and decoder, allows the model to have a much powerful transformation capability. The major difference is that the Jacobian matrix based penalty term is given in a layer-wise manner instead of end-to-end to reduce the computation complexity.

		\subsubsection{Test-time Adversarial Detection}
		\label{sec:adv-detect}
		
		A similar idea to the denoising approach (therefore may have overlap with the denoising approach) is by creating a detector in addition to the original classifier. Nguyen et al. \cite{nguyen2015deep} proposes that the detector can be simply implemented by adding a new {\em adversarial class} in the last layer of the original neural network. This idea is explored by Grosse et al. \cite{grosse2017on}, who show that adversarial examples are not drawn from the same distribution as the normal training data, and can be detected by statistical tests. 
		Instead of adding an extra adversarial class as is implemented by \cite{grosse2017on}, Metzen et al. \cite{metzen2017on} proposed to create a subnetwork which is trained on the binary classification task of distinguishing normal data from data adversarial inputs. Another benefits of this approach is that the detector is trained entirely independent of the original DNN, without affecting the classification accuracy on the legitimate inputs. 
		
		One interesting idea of adversarial detection is based on the intuition that unnecessarily large feature input spaces that are unimportant to many of the classification tasks, is one of the main reasons that leaves a back-door to the possibility of adversarial construction. Carlini et al. \cite{carlini2016hidden} notice that lowering the sampling rate can help to defend against the adversarial voice commands. Later, Xu et al. \cite{xu2018feature} propose a strategy to reduce the degrees of freedom available by ``squeezing'' out unnecessary input features, namely the ``feature squeezing''.
		More concretely, they explore two types of squeezing, including reducing the color depth of images, and using local or non-local smoothing (or blurring) to reduce the variation among pixels. 
		Their empirical study prove that these squeezing method affect very little on the classifier accuracy, but can largely remove the impact of the adversarial perturbation. Thus, the test-time adversarial detector can be built by comparing the model's prediction on the received input with its ``squeezed version'' to see if the outputs of the two versions are significantly different.  
		
		After studying a number of publications, Carlini and Wagner \cite{carlini2017adversarial} conclude with a number of common techniques involved in the detection methods, including: 1) the use of a second neural network for dedicated to classify if an input is adversarial; 2) the use of PCA to detect statistical properties of the input image or network parameters; 3) statistical tests to detect the difference of distribution between the natural images and adversarial inputs, and 4) performing input normalization with randomization and blurring. 
		
		\subsubsection{MagNet}
		{\em MagNet} is a popular defending mechanism that combines the idea of adversarial detection and distortion removal. From the perspective of representation learning, an autoencoder is capable of learning a low-dimensional manifold (e.g., 10 or 50 dimensions) that embeds in the high-dimensional input space (e.g., $28 \times 28 = 784$ dimensions for MNIST dataset), such that only points lie on the manifold are sensible. Therefore, the latent codes of the data points that follow the same generation process of the unperturbed training data should lie on the same manifold of that of the training data, and thus have low reconstruction errors. Based on this intuition, the {\em MagNet} method proposed by Meng and Chen \cite{meng2017magnet} separate the defending strategy into two components: {\em detector} defense and {\em reformer} defense, both based on the auto-encoders. 
		
		A {\em detector} is designed to ensure that the reconstruction error for an input $\| \mathrm{ae}(\boldsymbol{x}) - \boldsymbol{x} \|_p$ can never be above a certain threshold $t_\text{re}$ derived from the validation set, where $\mathrm{ae}(\cdot)$ denotes the reconstruction function of the autoencoder. Otherwise, this input is considered to be an adversarial input that will be rejected by the defender. Inputs that have low re-construction error are not necessarily close to the legitimate manifold. Therefore, one can also use the Jensen-Shannon divergence between the distributions that generated by the softmax function of the classifier on both the original input $f(\boldsymbol{x})$ and the re-constructed input $f(\mathrm{ae}(\boldsymbol{x}))$ as an extension, to estimate how likely the input is an adversarial example. 
		
		A {\em reformer} behave much similar to the DCN approach by Gu and Rigazio \cite{gu2014towards}, which tries to move the input point towards the manifold of normal examples by using auto-encoder, such that the adversarial perturbation can be removed in the newly reconstructed input. MagNet further strengthen the robustness of the reformer defense by building not just one, but a collection of auto-encoders, and one reformer network is chosen randomly for a new input example at test time.
		
		% By comparing to a previous adversarial detection method proposed by Metzen et al. \cite {metzen2017detecting}, MagNet has the advantage that it does not require any training data on the adversarial examples. 
		
		\subsubsection{Defense-GAN}
		\label{sec:gan-defend}
		
		In Section \ref{sec:gan-attack}, we introduced an attacking method based on the generative adversarial network, which constructs the input perturbation based on the latent space $\boldsymbol{z} \in \mathcal{Z}$. Conversely, an analogous GAN based defending method, namely the Defense-GAN, is proposed by Samangouei et al. \cite{samangouei2018defense}, 
		which leverages the expressive capability of generative models to defend deep neural networks against such attacks. 
		 
		A properly trained generator $\mathcal{G}$ from GAN (especially the W-GAN \cite{arjovsky2017wasserstein}) is capable of modeling the distribution of unperturbed training images $\mathbb{P}_\text{data}$. Formally speaking, for input data $\boldsymbol{x}$ that follow the distribution of unperturbed image, the reconstruction error of the generator converges to zeros: 
		\begin{equation}
			\mathbb{E}_{\boldsymbol{x} \sim \mathbb{P}_\text{data}} \left[ \min_{\boldsymbol{z}} \| \mathcal{G}_t{\boldsymbol{z}} - \boldsymbol{x} \|_2  \right] \to 0 \text{,}
		\end{equation}
		where $\mathcal{G}_t$ is the generator $\mathcal{G}$ after $t$ steps of the training \cite{kabkab2018task}. 
		 
\begin{figure}
		\center
			\includegraphics[width=0.45\textwidth]{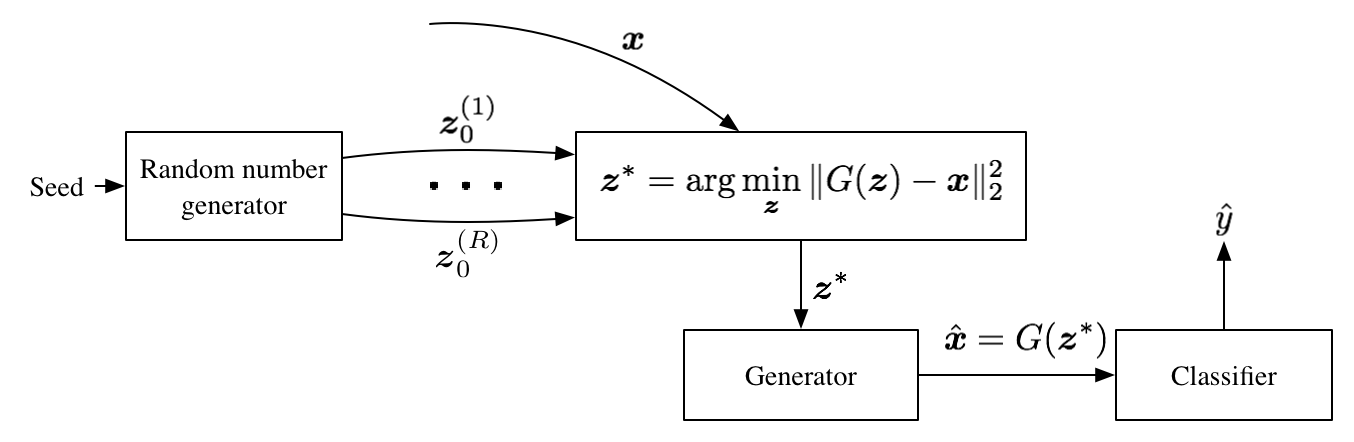}
			\caption{The overall working process of the Defense-GAN. Since finding the optimial $\boldsymbol{z}^*$ is non-convex, there are multiple starting points $\boldsymbol{z}_0^{(i)}$. }
			\label{fig:defense-gan}
		\end{figure}

		At test time, the algorithms first first map the new input $\boldsymbol{x}$ to the latent space by minimizing the reconstruction error:
		\begin{equation}
			\boldsymbol{z}^* = \argmin_{z} \| \mathcal{G}(\boldsymbol{z} - \boldsymbol{x}) \|_2^2
		\end{equation}
		by using common gradient descent optimization. Then, instead of the original input $\boldsymbol{x}$ (with potential adversarial perturbation), the classifier is with a newly generated example $\hat{\boldsymbol{x}}$. The whole process can be described by Figure \ref{fig:defense-gan}. 
		Since the generator $\mathcal{G}$ is expected to perfectly model the distribution of the unperturbed inputs, the reconstruction process can effectively reduce any potential adversarial noise.

		\subsection{The Fragility of Defenses}
		
		Despite the great efforts made by the community, the security guarantee concerning the adversarial attacks on general deep learning models is still inconclusive. Designing a ``universal'' defense approach seems to be impractical so far.
		A large number of existing defending techniques, though were claimed to be robust when proposed, have already been proven vulnerable. 
		For instance, 
		Carlini and Wagner \cite{carlini2017adversarial} studied a number of the detection-based adversarial learning techniques, 
		and come with the conclusion that the detection-based methods are not sufficiently strong to defend well design attacks, for both white-box or black-box attacks. 
		In addition, they also show that the once-important Defensive Distillation and MagNet defenses are either completely or mostly broken \cite{carlini2016defensive, carlini2017magnet}. What makes it worse is that, even combining multiple weak defense mechanisms do not promise a stronger one \cite{he2017adversarial}. 
 		
		Athalye et al. \cite{athalye2018obfuscated} studies this problem from the perspective of the model's gradient, which plays a central role in enhancing the training process. They prove that a wide range of defending strategies are based on the so called {\em obfuscated gradients}, which do not guarantee security against adversarial attacks as expected. They classify the obfuscated gradients into three sub-categories, and provide an effective attacking strategy for each: 
	\begin{itemize}
		\item {\bf Shattered Gradients} either due to the defense is non-differentiable, or causes the true gradient signal incorrect, and can be compromised by {\em Backward Pass Differentiable Approximation};
		\item {\bf Stochastic Gradients} caused by randomized defenses, either the network itself is randomized or the input is randomized, and can be compromised by {\em Differentiating over Randomness};
		\item {\bf Exploding or Vanishing Gradients} often caused by defenses that consist of multiple iterations of neural network evaluation, and can be compromised by {\em Reparameterization}.
	\end{itemize}
		They empirically show that 7 out of 8 defending strategies appear in ICLR 2018 conferences fall into these categories, all of which can be compromised by their attacking strategies. 
		
		The current empirical results force us to give more careful considerations on this question: Does the absolute security really exist? 
		Our pursue of more powerful attacking and defending strategies may form an infinite loop, which can be endless  after all.

\section{Extended Studies}

	\subsection{Verifying the Robustness}

	Absolute security is not possible by the state-of-the-art though, it is often desired to have a clear view of the measurable risk, when a system is deployed, and leads to a new research topic of {\em neural network verification}. 
	The key challenge of verifying the robustness of a system lies in the need to generalize the known valid inputs to a new dataset, whose size is often much larger than the {\em test set} for most benchmarks. 
	However, the asymmetry between the two sides is that when the attacking side needs only to demonstrate one way to succeed, the defending side must prove the in-existence of such possibility \cite{athalye2018obfuscated}, when it is not possible to exhaustively enumerate all valid input points near which the classifier is expected to give a approximately constant output. 
	Therefore, the verification problem can be reduced to finding or disprove the existence of adversarial examples. 
	
	Carlini et al. suggests that there are two general approaches to evaluate the robustness of a learning algorithm \cite{carlini2017towards}: 
	\begin{enumerate}
		\item When a theoretical proof is attainable, a lower-bound of the robustness can be obtained, which tells the maximum risk that the system may face; 
		\item Or, an upper-bound by constructing a successful attack can be obtained, to indicate the minimum risk of the system.
	\end{enumerate} 
	Intuitively, the theoretical lower-bound is more interesting to the application, but turns out much more difficult to obtain. In addition, making assumptions or approximations are always necessary when studying the theoretical proofs for deep neural networks, thus the claimed guarantee may not be the actual guarantee when these assumptions or approximations no longer holds.  
	
	Research works on this topic starts much earlier than adversarial learning in today's deep learning sense, which can be traced back to Pulina et al. \cite{pulina2010abstraction, pulina2012challenging}, who was looking for a method to ensure the body position would stay in the safe region when controlled by a neural network. 
	They developed one of the earliest verification system for Multi-Layer Perceptrons (MLPs) by approximating the sigmoid activation function thus is reducible to constrained SAT problem, and can be solved by Satisfiability Modulo Theories (SMT) solvers \cite{barrett2009satisfiability}.  
	Though they considers only a very small dimension input space with very limited size of training data, the idea of employing SMT solvers in this research topic has considerable influence on future works. 
	
	In the deep learning area, the Clever-Hans project by Papernot et al. \cite{papernot2016cleverhans,papernot2018cleverhans} is one of the most important projects of this type\footnote{https://github.com/tensorflow/cleverhans}, which provides a software library to standardized reference implementations of both adversarial construction and adversarial training methods. They advocate the concept of {\em verification} to replace the common {\em testing} method to study the robustness of deep neural networks in adversarial environments. 
	Similarly, Huang et al. \cite{huang2017safety} propose a Deep Learning Verification framework (DLV) that enables exhaustive search of the region partially based on 
	But they feature their work by restricting the verification task on a set of small regions, and obtain the strength to guarantee that adversarial examples must be found if any of them exist for that given region and family of manipulations. 
	
	Some researches on the other side, impose some restrictions on the type of classifiers to get more interesting results. 
	For instance, Fawzi et al. \cite{fawzi2018analysis} study specifically the linear and quadratic family of classifiers, and proved theoretically the fundamental upper and lower bounds of the robustness against adversarial perturbation.
	Katz et al. \cite{katz2017reluplex} suggest that the non-linear activation function is the main reason for the difficulty of studying the properties of deep neural networks. As such, they focus specifically on the Rectified Linear Unit (ReLU) activation function \cite{glorot2011deep, jarrett2009best, nair2010rectified}, and proposed the Reluplex (ReLU with Simplex), another verification system that works much more efficient by using an extended version of the simplex method.

	\subsection{Adversarial in Reinforcement Learning}
	
	Reinforcement learning (RL) is a special type of machine learning tasks, in which an intelligent agent are required to learn how best to interact with the environment to maximize the accumulated or expected rewards $R$ it is expected to receive. 
	At each time step $t$, the agent is able to observe the environment $\mathcal{E}$ at some state {s}, and decide its action $a_t$. When the environment receives the action $a_t$, its state is transferred into a new state $s_{t+1}$ with a probability $P(s_{t+1} | s_t, a_t)$ conditioned on $s$ and $a_t$. 
	Reinforcement learning differs from the usual supervised learning task in that the training data are not prepared in advance, but requires the agent to collect through its experience with the environment.
	However, both class of machine learning algorithms are fundamentally based on function fitting. From the RL's perspective, 	

	Huang et al. \cite{huang2017adversarial} firstly study the problem of adversarial attacks in reinforcement learning settings, and conclude that such attacks are also effective when targeting neural network policies. 	 
	They experiment with three major class of RL learning algorithms: A3C (Asynchronous Advance Actor-Criticor) \cite{mnih2016asynchronous}, TRPO (Trusted Region Policy Optimization) \cite{schulman2015trust}, and DQN (Deep Q-Networks) \cite{mnih2013playing}, and show that just like in the supervised case, even in the presence of perturbations perceivable to human may trick the agent to choose wrong actions and reduce the overall return.

	Lin et al. \cite{lin2017tactics} extend the research by Huang et al. by introducing the {\em strategically-timed attack} and {\em enchanting attack}.
	In strategically-timed attack, the adversarial choose to only attack the critical decision time-point by strategically selects a subset of time steps instead of attacking at every time step. The intuition behind the criterion of being critical is that when a well-trained policy shows a strong preference of choosing a certain action, it is often means choosing other options may cause a drastic reduction of the future rewards. This can be easily captured by a relative action preference function $c$ with e threshold parameter $\beta$.
	Instead of simply reducing the overall return, enchanting attack's goal however, is to trick the agent to enter a targeted state $s_g$ from current state $s_t$ of step $t$ after $H$ steps, by creating a sequence of adversarial examples $\{ s_{t+1} + \delta{t+1} , \dots , s_{t+H} + \delta{t+H} \}$. 
	This requires a combined effort of a generative model to predict the future states, and a planning model to generate a preferred sequence of actions. 

	\subsection{Attacks in Physical World}
	
	Verifying the robustness of the models became firstly interested to physical control systems as they are  machine learning algorithms due to their huge stack of potential loss \cite{pulina2010abstraction, katz2017reluplex}.
	However, as deep learning models are more involved in physical-world systems like auto-driving, this issue began to draw attention of both communities.
	Manipulating images in their digital format is comparatively simple. But when image inputs are obtained directly from a digital camera sensor, when alternating the inputs from the physical world may face various restrictions. 
	As remarked by \cite{lu2017no}, that creating adversarial attacks in the physical world can have a lot difference from that in the digital world. Many of the known methods become less effective due to reasons like the change of distance or the lighting condition etc. 

	Despite these difficulties, researchers have demonstrated such attacks in realistic scenarios  \cite{papernot2016transferability, papernot2016practical, kurakin2016adversarial_physical}. 
	Such attacks may also exists for inputs other than the images. For instance, Carlini et al.\cite{carlini2016hidden} show that audio signals can be synthesized in a way that is meaningless to human but may be recognized by smart phones as voice commands. 
	Papernot et al. \cite{papernot2017practical} point out that, attackers could target autonomous vehicles by using stickers or paint to create an adversarial stop sign, that the vehicle would interpret as some other sign like {\em yield}.
	There have been works to develop verification platforms focus exclusively on the potential adversarial risks of deep learning models exclusively to simulate physical world environment. For instance, by employing the idea of adversarial learning, Pei et al. \cite{DBLP:conf/sosp/PeiCYJ17} create a self-driving test platform called {\em DeepXplore} to enhance the robustness of the self-driving systems.

\section{Conclusions}
	
	Adversarial learning is a newly emerged topic in the deep learning community, which however has a relatively long history in the information security area. 
	The interpretation of the term ``adversarial'' in the deep learning community implies a restriction over the attacking method, that is adding an imperceivable amount of perturbation to the original input to influence the final output of the classifier. 
	 
	In this paper, we present a selective review on attacking and defending strategies, which only cover a tiny fractions of what have been proposed so far. 
	Whereas the mini-max game seems to be endless loop, in which new methods and approaches of both sides emerge constantly. 
	Studying such phenomenon does not only provide us with a tool of verifying the robustness and security of deep learning based system, but also an insight of the behavior of deep neural networks. 
	Besides, it also leads to several new research topics, among which the Generative Adversarial Networks (GAN), a generative model that takes advantage of a mini-max game with a discriminator to find the update gradients of its parameters, may be the most influential one. 
	This paper also talks briefly about the adversarial attacks in more complex scenarios, such as the reinforcement learning, or even the physical world attack. 
    
	One obvious feature of the current adversarial deep learning is that most of the researches now focus exclusively on the computer vision problem, especially the image classification problem, which is heavily dependent on the convolution network architecture. 
	There are drawbacks that known to be inherent in the convolution network architecture, 
	for instance, by using the sub-sampling to keep translation invariance, the precise spatial relationships between higher-level parts are lost. 
    Therefore, we expect more studies on the non-convolution architecture to appear in future researches.

\bibliographystyle{unsrt}
\bibliography{main}

\end{document}